%% file: atemplate.tex
\documentclass[a4paper]{IEEEtran}
\pdfoutput=1

\usepackage{amsmath}
\usepackage{cleveref}
\usepackage{graphicx}
\usepackage{url}
\usepackage[square,numbers]{natbib}
\usepackage{etoolbox}
\usepackage{booktabs}

\usepackage{mathtools}
\usepackage{siunitx}
\usepackage[caption=false, font=footnotesize]{subfig}

\DeclareMathOperator*{\argmin}{arg\,min}

\title{Motion-based extrinsic sensor-to-sensor calibration: Effect of reference frame selection for new and existing methods}

\author{Tuomas Välimäki $^{1,}$*, Bharath Garigipati $^{1}$ and Reza Ghabcheloo $^{1}$\thanks{$^{1}$ Automation Technology and Mechanical Engineering, Tampere University, Tampere, Finland\\ * Correspondence: tuomas.valimaki@tuni.fi}}

\begin{document}
\maketitle
\thispagestyle{empty}
\pagestyle{empty}

\crefname{figure}{Figure}{Figures}
\Crefname{figure}{Figure}{Figures}
\crefname{table}{Table}{Tables}
\Crefname{table}{Table}{Tables}
\crefname{equation}{}{}
\Crefname{equation}{Equation}{Equations}

\sisetup{detect-all = true}
\robustify\bfseries

\begin{abstract}
This paper studies the effect of reference frame selection in sensor-to-sensor extrinsic calibration when formulated as a motion-based hand-eye calibration problem. Different reference selection options are tested under varying noise conditions in simulation, and the findings are validated with real data from the KITTI dataset. We propose two nonlinear cost functions for optimization and compare them with four state-of-the-art methods. One of the proposed cost functions incorporates outlier rejection to improve calibration performance and was shown to significantly improve performance in the presence of outliers, and either match or outperform the other algorithms in other noise conditions. However, the performance gain from reference frame selection was deemed larger than that from algorithm selection. In addition, we show that with realistic noise, the reference frame selection method commonly used in literature is inferior to other tested options, and that relative error metrics are not reliable for telling which method achieves best calibration performance.
\end{abstract}


\section{Introduction}

With growing industry interest in automating many tasks performed with mobile working machines such as forklifts, excavators, or harvesters, easy-to-use sensor to sensor calibration is essential in enabling sensor fusion for the desired algorithms and functionality. Traditional sensor calibration methods rely on structured environments and/or calibration targets, meaning laborious setups for calibration data collection. Multiple targetless sensor to sensor calibration methods have been proposed to circumvent these issues, but most still require a shared field of view for sensors, which is not always possible or desirable, and limits the types of sensors that can be used.

Recently, many motion-based extrinsic calibration methods, that treat the process as a hand-eye calibration problem, have been proposed. Formulated in this manner, the calibration i) does not require calibrated targets, ii) does not require a shared field of view, and iii) can work with multimodal sensors, requiring only the sensor trajectories. These properties make the hand-eye calibration approach especially suitable for mobile working machines that often incorporate multimodal sensor arrays that do not have overlapping fields of view. When no calibration targets are needed, the recalibration of the sensor array is also possible in field conditions, which is very important for large machines that cannot be easily taken back to shop for repairs.

In the traditional hand-eye calibration problem, with a camera mounted on a robotic arm, there is very little noise when compared with sensor trajectories generated using SLAM (Simultaneous Localization and Mapping) algorithms. What noise there is, is best modeled as additive Gaussian noise. SLAM trajectories, however, may contain noticeable drift and outliers in addition to general noise. To the authors best knowledge, no existing literature studies how the effect of these different types of noise on the calibration performance could be mitigated through the selection of reference frames.

In this paper we compare multiple state-of-the-art motion-based extrinsic calibration algorithms and data preprocessing under varying noise conditions. The first novelty comes from considering the effect of reference frame selection in calibration performance. In addition, we propose a nonlinear optimization method with outlier rejection that is shown to improve performance under some of the conditions. We show that the reference frame selection method commonly used in literature is inferior in presence of noise typical for SLAM generated trajectories. In practice, since the ground truth is unknown, the only option to compare two methods is to compare which one reduces the relative error the most. However, using simulated data with known ground truth information, we show that the relative error metrics are not reliable for telling which method is better. The findings are demonstrated both using a large set of simulated data and the publicly available KITTI dataset~\citep{Geiger2013}. We also provide general guidelines for reference frame selection in data preprocessing and propose directions for future research.

\begin{figure}[t]
    \centering
    \includegraphics[width=0.85\linewidth]{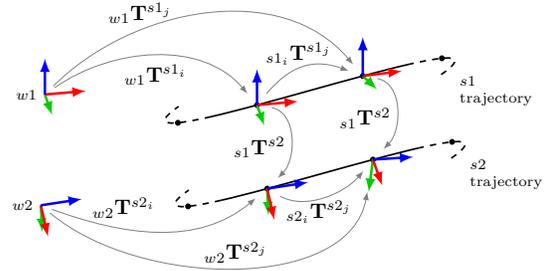}
    \caption{Graphical representation of the motion-based sensor to sensor calibration concept, where pose $i$ is used as the reference frame for pose $j$.}%
    \label{fig:hand-eye}
\end{figure}

\section{Related work}

Motion-based extrinsic sensor to sensor calibration has a close affinity to hand-eye calibration from manipulator literature, and many of the same algorithms apply. There are two main types of algorithms that solve the hand-eye calibration problem, \emph{separable} algorithms that solve the orientation and translation parts in separate linear stages, and \emph{simultaneous} algorithms that solve the complete nonlinear problem in a single stage. \emph{Separable} solutions are generally simpler but more error-prone, since the errors from rotation estimation propagate to the translational part. Lately, also other methods to solve the hand-eye calibration have been proposed, such as factor graph optimization~\citep{Koide2019a} and even neural networks~\citep{Pachtrachai2021}.

In \citep{Taylor2015} the authors present a \emph{separable} motion based calibration method for multimodal sensors. A coarse alignment is based on a modified Kabsch algorithm, that is then refined using sensor specific metrics for sensors with overlapping field of view. The method is later improved in~\citep{Taylor2016}. Similarly, a lidar-to-camera calibration method based on trajectory matching for coarse alignment and a refinement step based on image features is presented in \citep{Park2020}. The \emph{separable} course alignment is applicable regardless of sensor type, but finetuning is limited to cameras.

In \citep{Zhuang1994}, an early work on \emph{simultaneous} solution, the problem is formulated using quaternions, and observability conditions are provided based on a new identification Jacobian. More recently, \citep{Pedrosa2021} presents a \emph{simultaneous} general approach for hand-eye calibration for multiple cameras, based on optimization of atomic transformations dubbed ATOM. In \citep{Aguiar2021}, the authors build upon ATOM to allow multimodal sensors, but the method still necessitates a camera due to reprojection error being used as optimization metric. Both calibration methods are target based and the sensors must share field of view.

A survey of state-of-the-art hand-eye calibration methods in the traditional robotic arm use case is presented in \citep{Ali2019}. The authors also present their own \emph{simultaneous} optimization method and propose multiple possible cost functions. The \emph{simultaneous} nonlinear optimization method proposed in our paper is similar to the formulation of Xc1 in \citep{Ali2019}, but differs in using rotation vectors instead of quaternions. The MATLAB implementation provided by the authors of \citep{Ali2019}, however, does not effectively include rotational errors at all in the cost function, but instead the cost for the rotational error is the norm of a unit quaternion. Our reimplementation replicates the behaviour implemented by the original authors. Our paper also proposes a novel modification to the cost function to add outlier rejection without resorting to a computationally heavy RANSAC~\citep{Fischler1981} framework.

Iterative \emph{simultaneous} optimization methods are shown to perform better than closed-form solutions, which have difficulties handling noisy or inconsistent data, in the traditional robotic arm use case~\citep{Pedrosa2021,Ali2019,Tabb2017}. However, to the authors best knowledge, no previous work discusses the data preprocessing choices, especially reference selection, and their effect on calibration performance under different noise conditions.

\section{Method}

\subsection{Calibration}%
\label{sec:calibration}

We introduce a notation, where we represent homogeneous transformations as $\mathbf{T}$, indicating the source and target frames as sub- and super-indices respectfully, e.g. $_{w1}\mathbf{T}^{s1_i}$ represents the transformation from the world frame associated with sensor 1 to the sensor pose at time instance $i$. The corresponding rotation and translation parts are referred to as $\{\mathbf{R},t\}$ respectfully, using the same sub- and super-indexing.

Consider two sensors, $s1$ and $s2$, installed on the same rigid body. When the body moves, the sensor trajectories are recorded e.g. using suitable SLAM algorithms. As presented in \cref{fig:hand-eye}, we can now form the relationship
\begin{equation}
    ({{}_{w1}\mathbf{T}^{s1_i}})^{-1} {}_{w1}\mathbf{T}^{s1_j} {}_{s1}\mathbf{T}^{s2} = {}_{s1}\mathbf{T}^{s2} ({{}_{w2}\mathbf{T}^{s2_i}})^{-1} {}_{w2}\mathbf{T}^{s2_j},
\end{equation}
or equivalently
\begin{equation}
    {}_{s1_i}\mathbf{T}^{s1_j} {}_{s1}\mathbf{T}^{s2} = {}_{s1}\mathbf{T}^{s2} {}_{s2_i}\mathbf{T}^{s2_j}, \label{eq:relation}
\end{equation}
which is the well known hand-eye calibration formulation $\mathbf{AX}=\mathbf{XB}$. Here, $\mathbf{A}$ and $\mathbf{B}$ comprise the relative transformations between two frames in the respective sensor trajectories and $\mathbf{X}$ is the unknown sensor to sensor transformation ${}_{s1}\mathbf{T}^{s2}$. Notice that the equality is valid for any pair of $(i,j)$ from the data, and we can solve the group of equations formed from the set of all selected pairs $S=\{(i,j),\dots\}$. The relative transformations are typically calculated between two consecutive poses, i.e. $i=j-1$, but this need not be the case as covered in Section~\ref{sec:data_processing}. We will also show that different selection methods result in more accurate calibration.

We analyzed several state-of-the-art calibration methods, namely those presented by \citet{Ali2019}, \citet{Park2020}, \citet{Taylor2015}, and \citet{Zhuang1994}. For \citep{Park2020} and \citep{Taylor2015}, only the coarse alignment is applicable, and from \citep{Ali2019} we use the cost Xc1. The algorithms are later referred to by only the first author. In addition, we propose our own nonlinear optimization method detailed in the following section.

\subsection{Proposed optimization}%
\label{sec:optimization}

We directly minimize the nonlinear relation in~\cref{eq:relation}. The unknowns for the optimization are $\{\theta, t\}$, where $\theta$ is the rotation vector corresponding with the axis-angle representation of ${}_{s1}\mathbf{R}^{s2}$, and the indices are dropped from ${}_{s1}t^{s2}$ for the sake of brevity.

We use two cost functions, in which
\begin{equation}
    \mathbf{M}_{i,j} \coloneqq
    {}_{s1_i}\mathbf{T}^{s1_j} [\theta,t]_T - [\theta,t]_T {}_{s2_i}\mathbf{T}^{s2_j}
    ,
\end{equation}
where the $[~]_T$ notation represents the homogeneous transformation formed from the parameters. The first cost is
\begin{equation}
    \{\theta,t\} =
    \argmin_{\theta,t} \sum_{(i,j)\in S}
    \left\lVert
    \mathbf{M}_{i,j}
    \right\rVert_2^2
    ,\label{eq:dnl}
\end{equation}
later referred to as direct non-linear (DNL), and the second one is
\begin{equation}
    \begin{split}
    \{\theta,t,\alpha\} = \argmin_{\theta,t,\alpha} \sum_{(i,j)\in S} \left( \alpha_{i,j}
    \left\lVert
    \mathbf{M}_{i,j}
    \right\rVert_2^2
    + (1-\alpha_{i,j}) c \right) \\
    \text{subject to:\quad} \begin{split} 0 &\le \alpha_{i,j} \le 1, \forall i,j\\ d &\le \sum_{i,j} \alpha_{i,j} \end{split},
    \end{split}\label{eq:dnlo}
\end{equation}
later referred to as direct non-linear with outlier rejection (DNLO). DNLO adds outlier rejection by incorporating a weighting vector $\alpha$ into the decision variables. The rejection can be tuned by appropriately selecting parameters $c$ and $d$. During optimization, if the error term $\left\lVert\mathbf{M}_{i,j}\right\rVert_2^2 > c$, the corresponding weight $\alpha_{i,j}$ will tend to zero and the error is regarded an outlier. Therefore, $c$ defines a threshold for error tolerance. Similarly, if $\left\lVert\mathbf{M}_{i,j}\right\rVert_2^2 < c$ then $\alpha_{i,j}$ will tend to one. The sum over $\alpha$ can therefore be regarded as the proportion of inliers, and $d$ defines the minimum proportion of weighted poses to keep. Our implementation uses Ipopt as part of CasADi~\citep{Andersson2019} for the minimization.

\subsection{Error metrics}%
\label{sec:error_metrics}

For evaluation purposes, we adopt the four error metrics used in~\citep{Ali2019}. The first two are derived from the relation~\cref{eq:relation}, the first being mean relative translation error
\begin{equation}
    \begin{split}
    e_{rt} = \frac{1}{\lvert S \rvert}\sum_{(i,j)\in S} \left\lVert
        {}_{s1_i}\mathbf{R}^{s1_j}{}_{s1}t^{s2} + {}_{s1_i}t^{s1_j} - \right.\\
        \left. {}_{s1}\mathbf{R}^{s2}{}_{s2_i}t^{s2_j} + {}_{s1}t^{s2}
    \right\rVert_2,
    \end{split}
\end{equation}
and the second being mean relative rotation error
\begin{equation}
    \begin{split}
    e_{rR} = \frac{1}{\lvert S \rvert}\sum_{(i,j)\in S} \left\lVert
        \big[ ({}_{s1}\mathbf{R}^{s2}{}_{s2_i}\mathbf{R}^{s2_j})^{-1}{}_{s1_i}\mathbf{R}^{s1_j}{}_{s1}\mathbf{R}^{s2} \big]_\theta
    \right\rVert_2,
    \end{split}
\end{equation}
where the notation $[~]_\theta$ represents the rotation vector formed from the rotation matrix and $\lvert S \rvert$ is the cardinality of set $S$.

The remaining two are defined as the difference from the known ground truth values $\{{}_{s1}\mathbf{R}^{s2}_{gt}, {}_{s1}t^{s2}_{gt}\}$, namely, the absolute translation error
\begin{equation}
    e_{at} = \left\lVert {}_{s1}t^{s2}_{gt} - {}_{s1}t^{s2} \right\rVert_2, \label{eq:eat}
\end{equation}
and the absolute rotation error
\begin{equation}
    e_{aR} = \left\lVert \big[{{}_{s1}\mathbf{R}^{s2}}^{-1}{}_{s1}\mathbf{R}^{s2}_{gt}\big]_\theta \right\rVert_2.
\end{equation}

\begin{figure}
\centering
\subfloat[]{\includegraphics[width=0.6\linewidth]{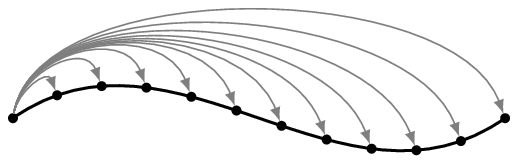}}\\
\subfloat[]{\includegraphics[width=0.6\linewidth]{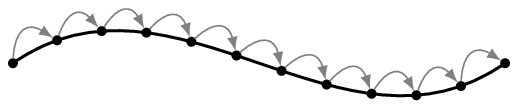}}\\
\subfloat[]{\includegraphics[width=0.6\linewidth]{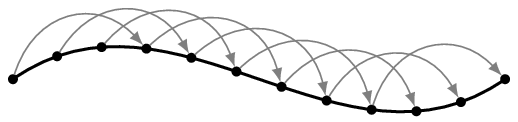}}\\
\subfloat[]{\includegraphics[width=0.6\linewidth]{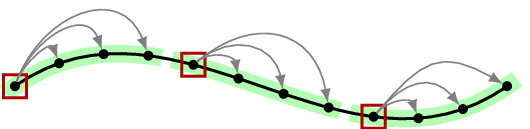}}
\caption{Different ways to define the reference frame for relative coordinates used in calibration. (\textbf{a}) $i=0$, (\textbf{b}) $i=j-1$, (\textbf{c}) $i=j-n$, and (\textbf{d}) $i=k_s$.}%
\label{fig:reference}
\end{figure}

\subsection{Reference selection}%
\label{sec:data_processing}

In addition to multiple calibration algorithms, we study different ways to select the pairs $(i,j)$ for the relative transformations needed in the hand-eye calibration problem introduced in Section~\ref{sec:calibration}. The studied reference frames for the relative coordinates are presented in \cref{fig:reference} and are namely
\begin{itemize}
    \item Case $A$: where all poses are w.r.t. the first pose of the trajectory, i.e. $i=0$. The method is expected to be particularly susceptible to drift due to the error constantly increasing the farther away the current pose is from the start of the trajectory.
    \item Case $B1$: where the reference for relative movement is the previous pose in trajectory, i.e. $i=j-1$. This is the \emph{de facto} reference used throughout previous studies. Keeping the transformations small reduces susceptibility to drift, but might prove vulnerable to noise. If the transformations between the poses are too small w.r.t. the noise, i.e. the signal-to-noise ratio worsens, the calibration performance likely suffers.
    \item Case $Bn$: a broader definition of the previous case where the reference is the $n$\textsuperscript{th} previous pose, i.e. $i=j-n$. Freedom in choosing $n$, provides a way to trade off between sensitivity to noise and drift. The difference to just keeping every $n$\textsuperscript{th} pose from the original trajectory is, that as opposed to drastically reducing the number of poses, only $n-1$ first poses are dropped.
    \item Case $Cn$: a keyframe based approach inspired by \citep{Schneider2019}, where the trajectory is divided into segments of equal length $n$. All poses in each segment are w.r.t. the first frame of the segment, the so called keyframe $k_s$, i.e. $i=k_s$. The method provides a mix of smaller and larger relative transformations that might prove useful in certain situations.
\end{itemize}

Note that there needs to be a minimum of 2 poses in the resulting relative transformations to solve for the unknowns in~\cref{eq:relation}, and there must be sufficient rotation and translation between the pairs $(i,j)$.

\begin{figure*}[t]
    \centering
    \subfloat[]{%
        \includegraphics[width=0.48\linewidth]{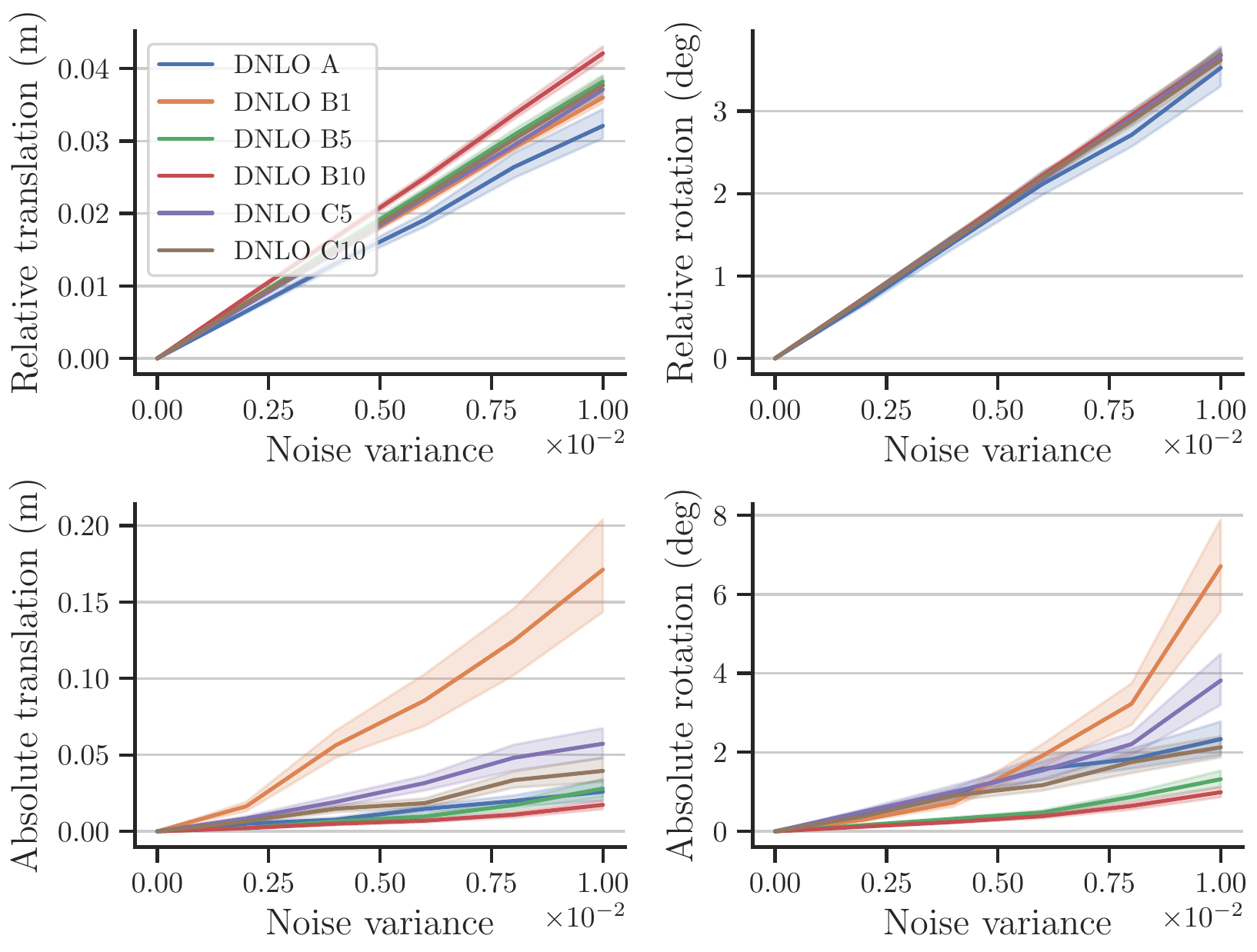}%
        \label{fig:dnlo_gaussian}%
    }\hfill
    \subfloat[]{%
        \includegraphics[width=0.48\linewidth]{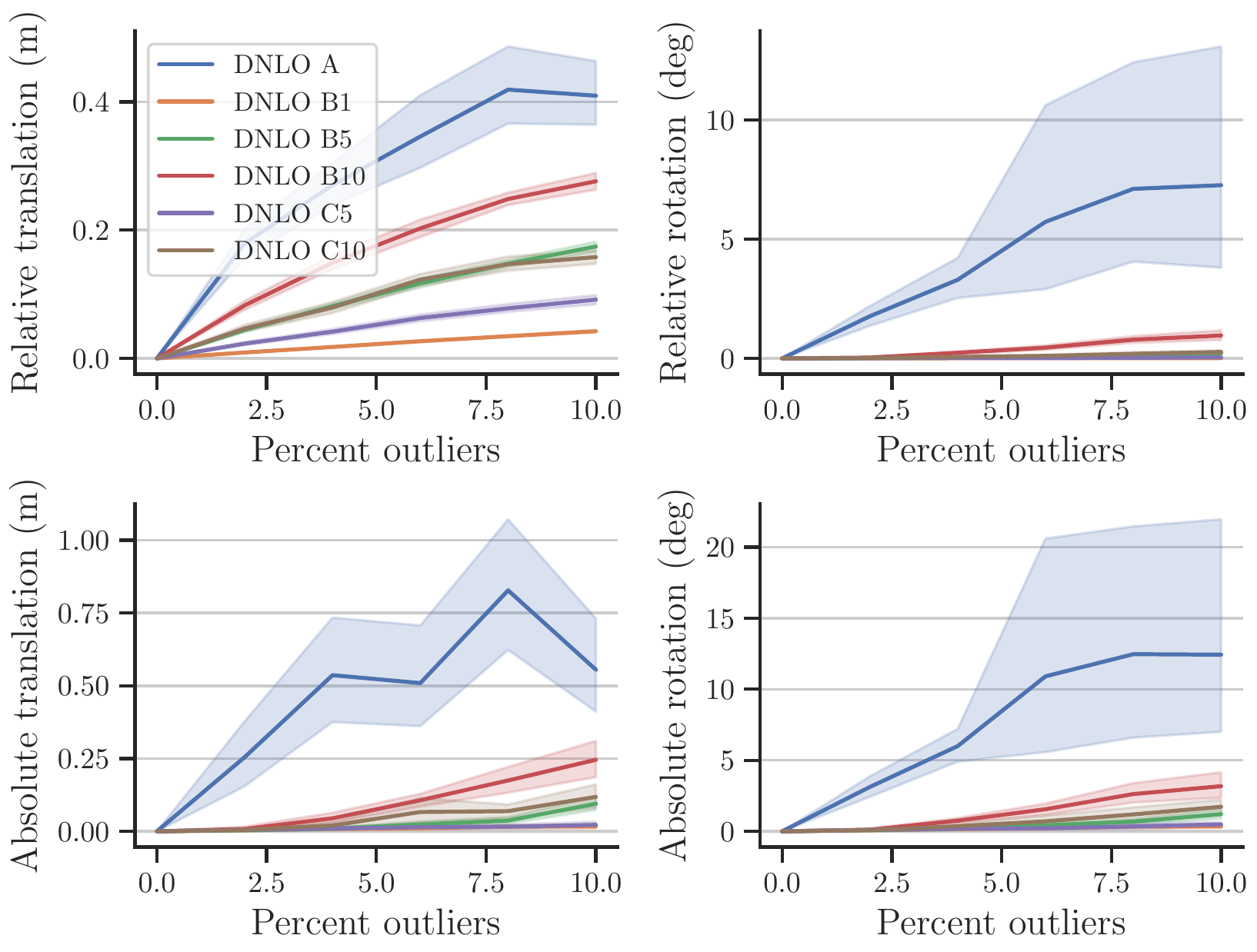}%
        \label{fig:dnlo_outliers}%
    }\\
    \subfloat[]{%
        \includegraphics[width=0.48\linewidth]{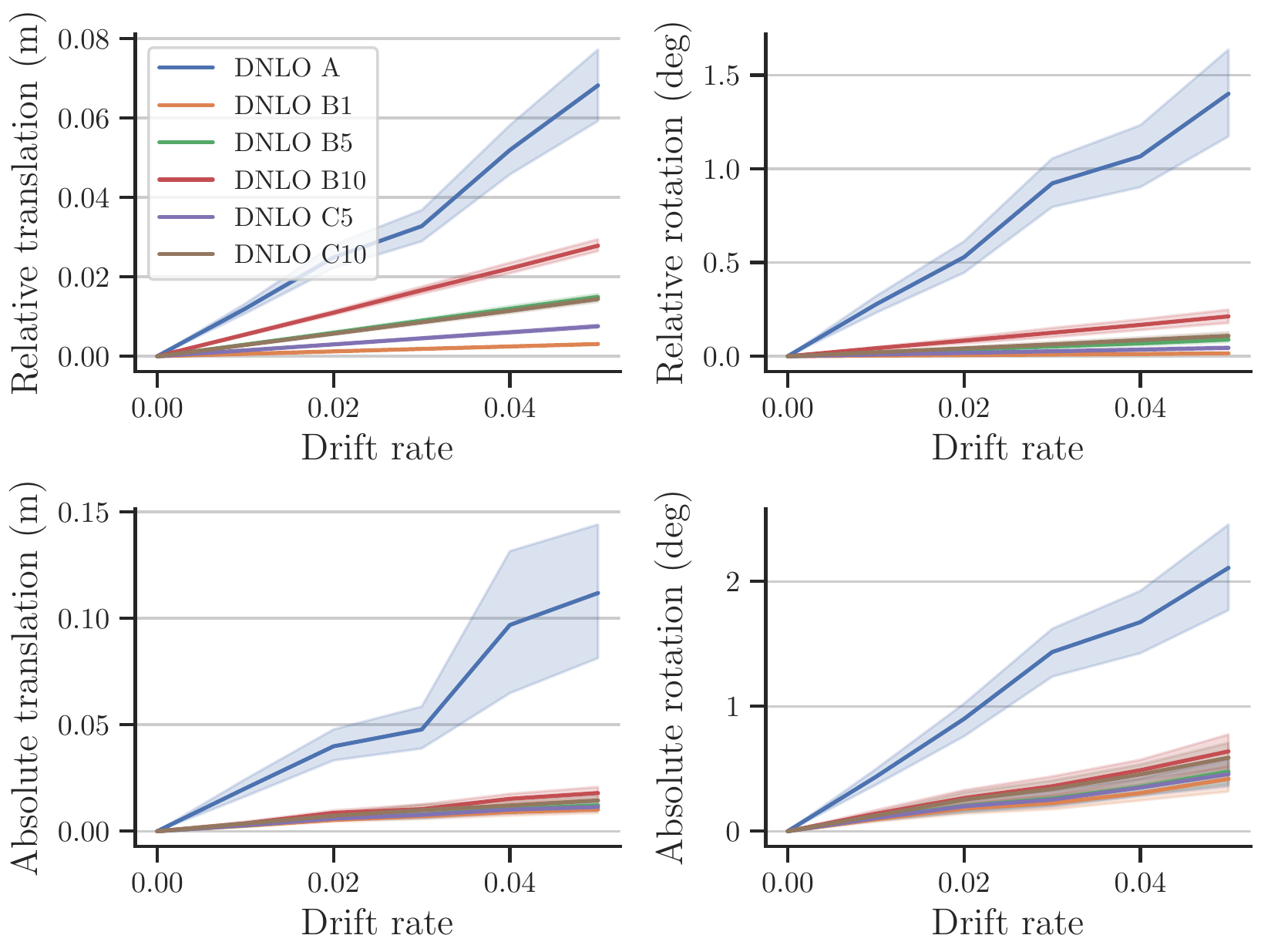}%
        \label{fig:dnlo_drift}%
    }\hfill
    \subfloat[]{%
        \includegraphics[width=0.48\linewidth]{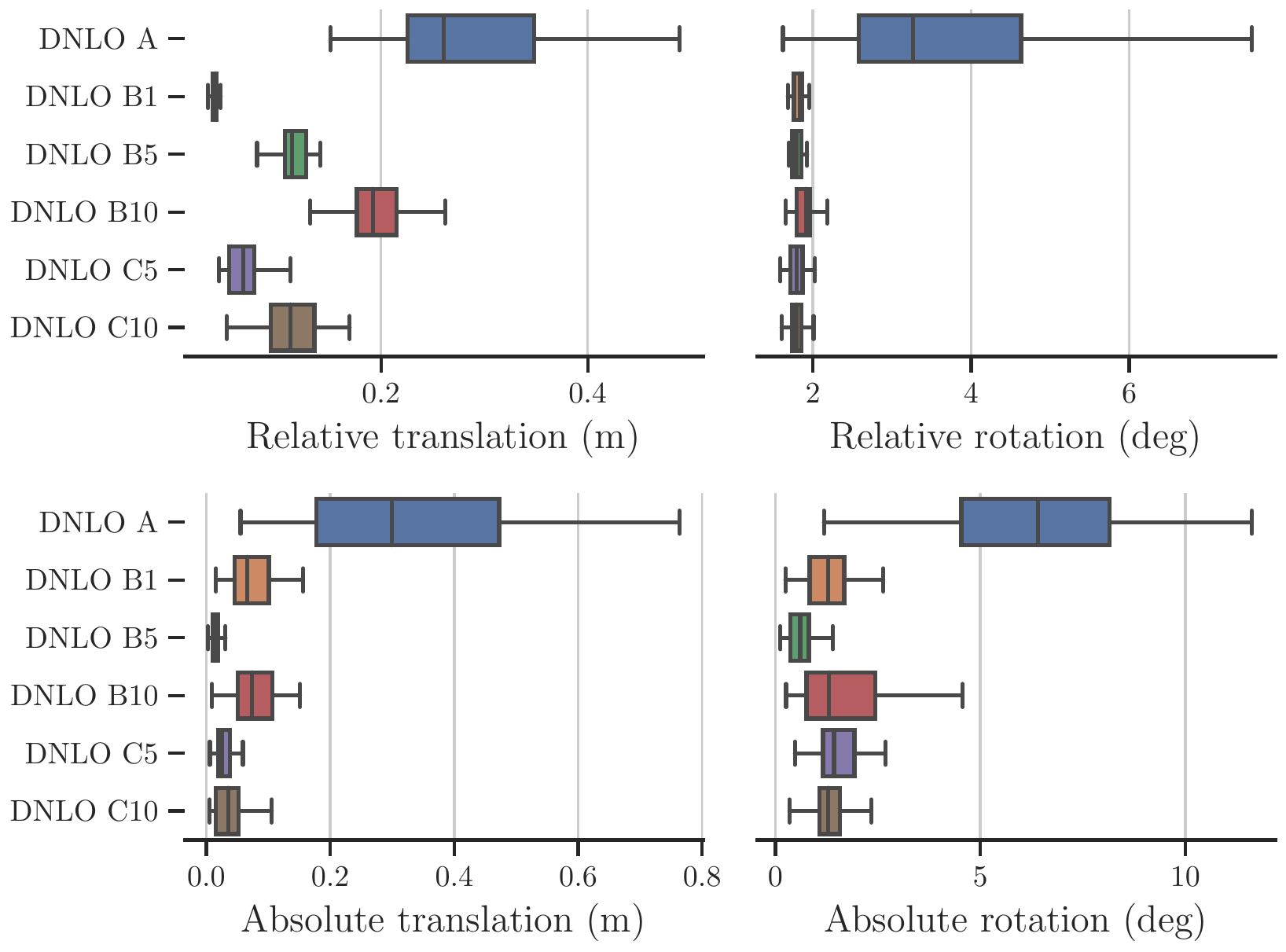}%
        \label{fig:dnlo_mixed}%
    }%
    \caption{Calibration errors on simulation data for all reference selection methods on DNLO: (\textbf{a}) added Gaussian noise, (\textbf{b}) added outliers, (\textbf{c}) added drift, and (\textbf{d}) mixed noise. Plots (\textbf{a}) through (\textbf{c}) display the mean and 95\% confidence intervals, whereas the boxplot (\textbf{d}) shows the median and quartiles.}%
    \label{fig:dnlo}
\end{figure*}

\begin{figure*}
\centering
\includegraphics[width=\linewidth]{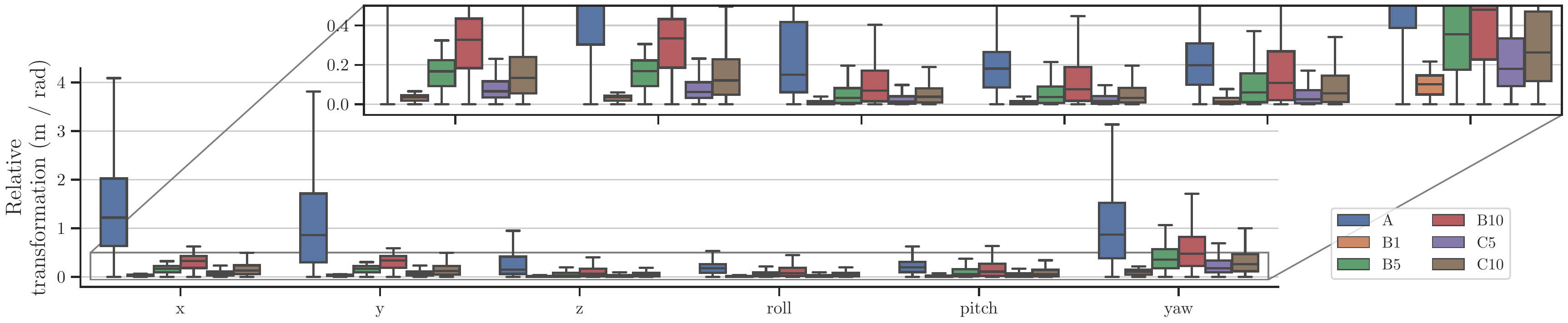}%
\caption{Median and quartiles of the absolute components of the resulting relative transformations when using the studied reference frame selection methods on the noiseless base trajectories of the simulation data.}%
\label{fig:transformation}
\end{figure*}

\begin{figure*}[t]
    \centering
    \subfloat[]{%
        \includegraphics[width=0.48\linewidth]{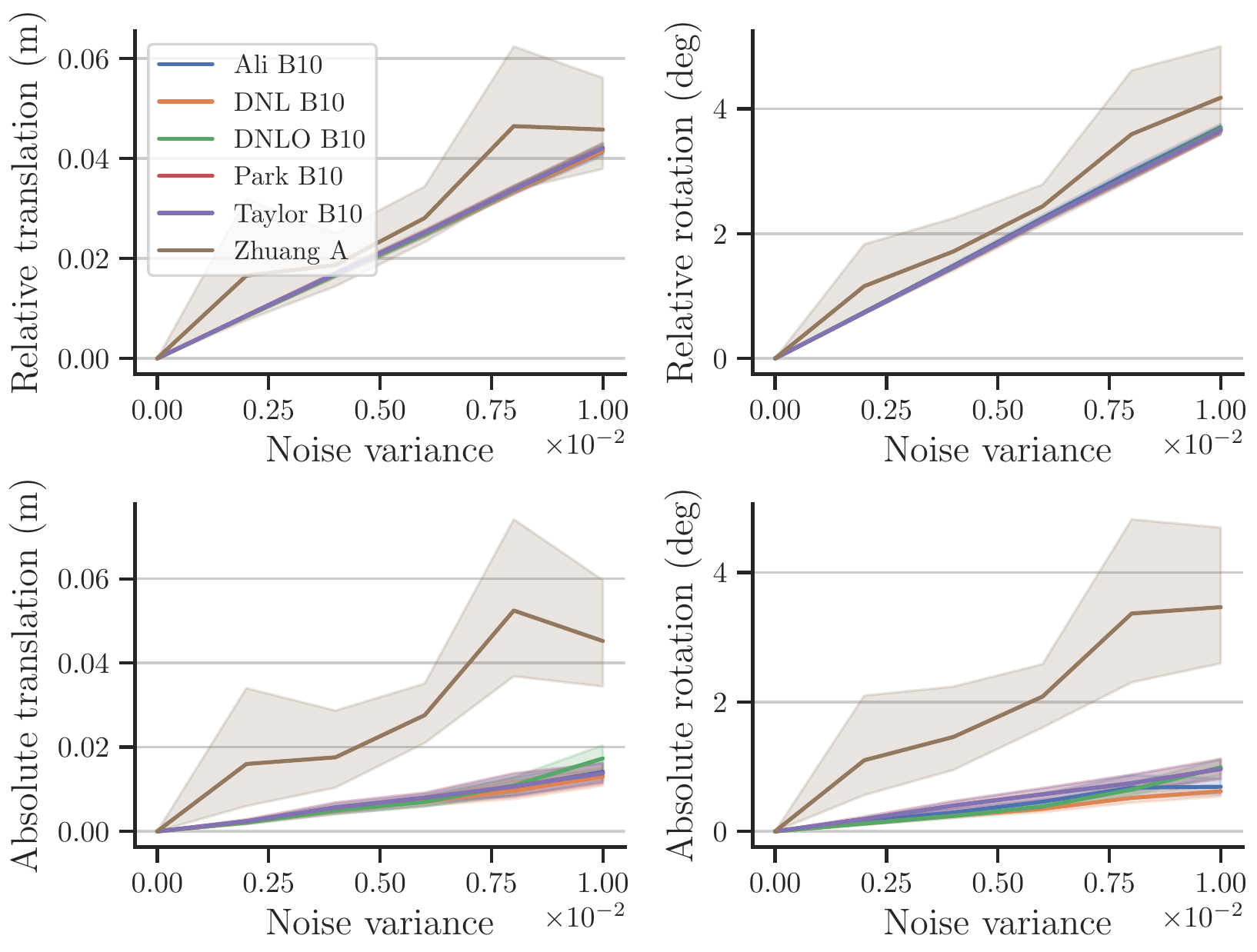}%
        \label{fig:gaussian}%
    }\hfill
    \subfloat[]{%
        \includegraphics[width=0.48\linewidth]{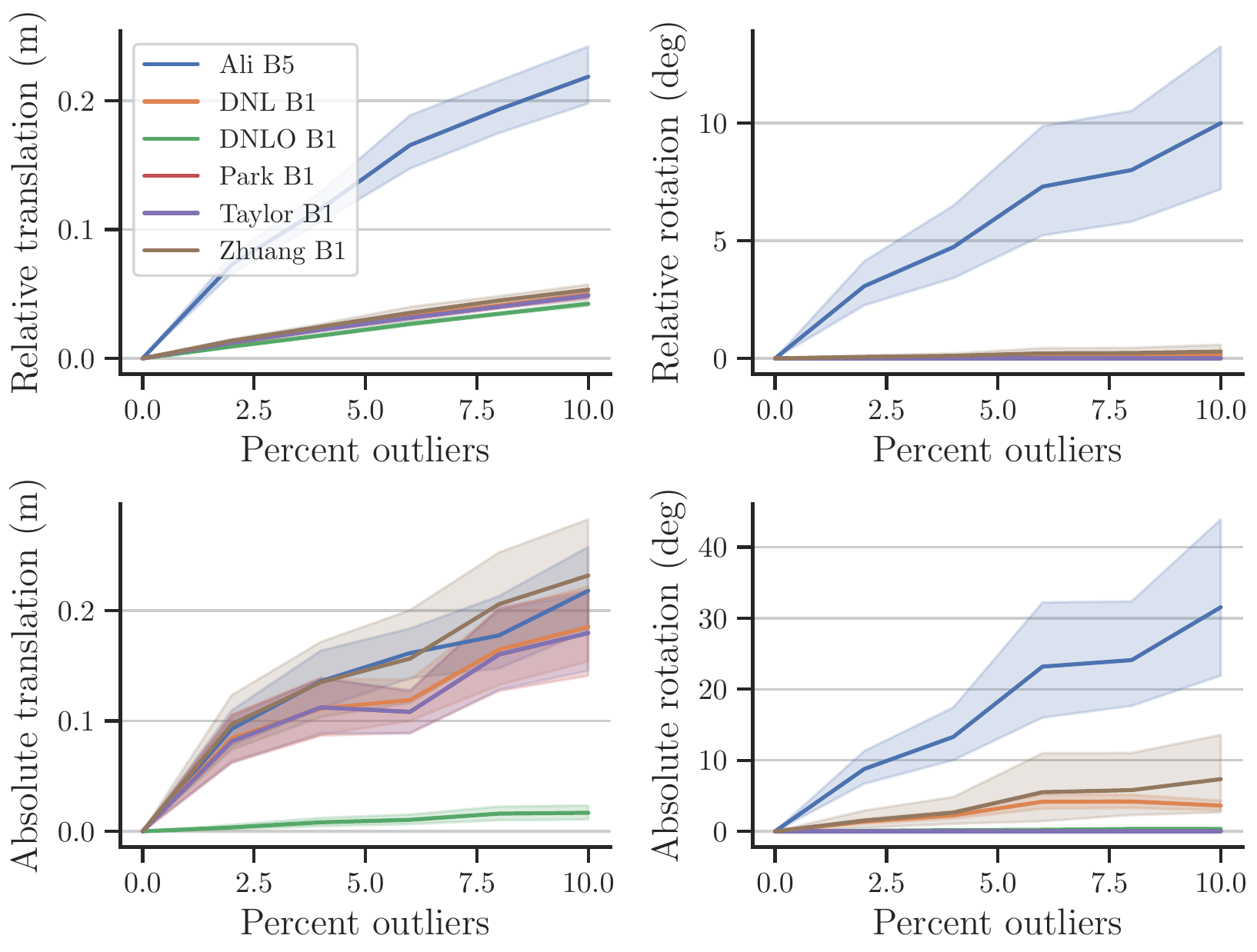}%
        \label{fig:outliers}%
    }\\
    \subfloat[]{%
        \includegraphics[width=0.48\linewidth]{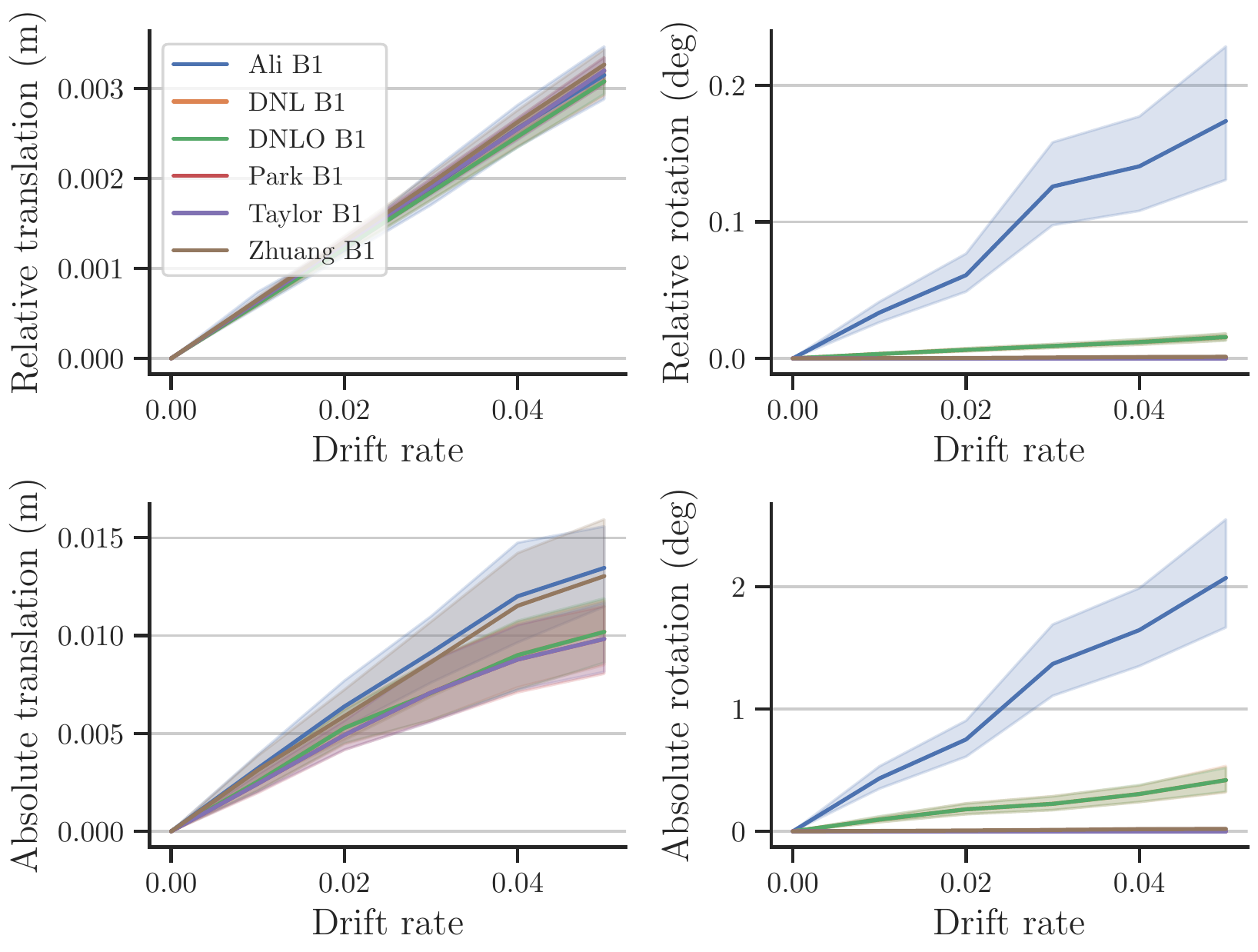}%
        \label{fig:drift}%
    }\hfill
    \subfloat[]{%
        \includegraphics[width=0.48\linewidth]{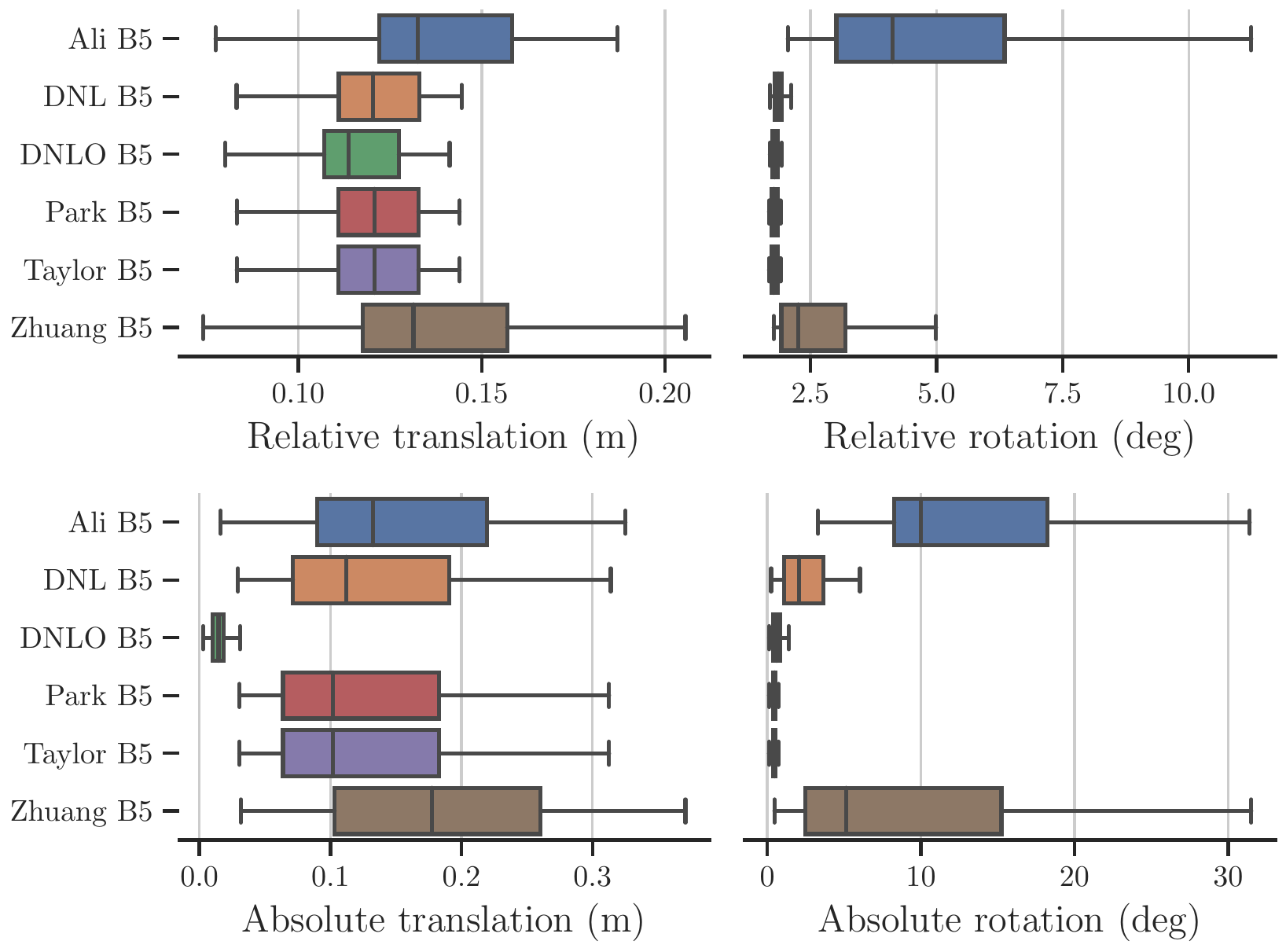}%
        \label{fig:mixed}%
    }%
    \caption{Calibration errors on simulation data for best performing variant of each algorithm: (\textbf{a}) added Gaussian noise, (\textbf{b}) added outliers, (\textbf{c}) added drift, and (\textbf{d}) mixed noise. Plots (\textbf{a}) through (\textbf{c}) display the mean and 95\% confidence intervals, whereas the boxplot (\textbf{d}) shows the median and quartiles.}%
    \label{fig:best}
\end{figure*}

\section{Simulation experiments}

\subsection{Data generation}%
\label{sec:data_generation}

For the purpose of simulation experiments, a random base trajectory is first generated for the primary sensor. The trajectory for the second sensor is then generated based on the sensor-to-sensor ground truth transformation ${}_{s1}\mathbf{T}^{s2}_{gt}$, after which both trajectories are subjected to different noise patterns to be expected in trajectories extracted using SLAM algorithms. The simulation experiments are based on 38 of such random trajectories.

The noise patterns used in the study were
\begin{itemize}
    \item Gaussian noise with varying $\sigma^2$, where the trajectories are right-multiplied with $[\tilde{\mathbf{R}}, \tilde{t}]_T$. The rotation matrix $\tilde{\mathbf{R}}$ is formed from Euler angles drawn from $\mathcal{N}(0,2\sigma^2)$ and the translation components in $\tilde{t}$ are drawn from $\mathcal{N}(0,\sigma^2)$.
    \item random jumps in $[x, y, z]$ -coordinates, where the translation components in $\tilde{t}$ are drawn from $\mathcal{N}(0,0.02)$. The variance 0.02 was selected because it is considerably larger than the noise levels in the previous point
    \item drift on randomly selected principal axis with increasing severity determined by \emph{drift rate}, i.e. how many meters per meter moved the error increases.
\end{itemize}
A range of each noise pattern is applied to all of the 38 generated base trajectories.

In addition to the above noise patterns individually, the algorithms were tested using a mixture of all three, which could be argued to be the most realistic case. For the mix of noises, the median values of the tested ranges were used for the noise components, namely $\sigma^2=\num{5e-3}$, 5\% outliers, and drift rate \SI{0.025}{m/m}. Usually, SLAM algorithm performance is reported using absolute trajectory error (ATE)~\citep{Sturm2012}. The average RMS~ATE for various vision based algorithms reported in~\citep{Campos2021} ranges from \SI{0.035}{m} to \SI{0.601}{m}. The average RMS~ATE for the mixed noise trajectories is \SI{0.364}{m}, so the selection of parameters can be considered reasonable. The authors did not manage to find a source for the typical composition of noise, and it is expected to vary based on the environment and sensors involved.

\subsection{Results}

We ran the calibration on the generated simulation data using all six calibration methods in the comparison. Each calibration method was tested using six different reference frame selection methods $A$, $B1$, $B5$, $B10$, $C5$, and $C10$, where the labels refer to the methods presented in Section~\ref{sec:data_processing}. The calibration results are compared using the error metrics presented in Section~\ref{sec:error_metrics}. Throughout the experiments, the parameters $c=0.01$ and $d=\frac{\lvert S \rvert}{2}$ were used for DNLO. The values were empirically chosen using the simulation data.

\subsubsection{Comparing reference selection methods}

Due to limited space to present the extensive results, we provide the comparison of all reference selection methods only for our DNLO in \cref{fig:dnlo}, and later compare the best performing variants for each algorithm. The extensive results for all methods and algorithms are provided as supplementary material. The graphs present the mean and 95\% confidence intervals of the calibration errors for all 38 trajectories, except for the mixed noise case, where we present the median and quartiles.

Firstly, it can be verified that without noise, the reference selection strategy makes no difference. Further, the results mostly follow expectations where longer reference distances are better with added noise, but worse with added outliers and drift. However, it is worth noticing that preprocessing option $A$, providing the largest transformations, is not the best option for even Gaussian noise, and noticeably worse for the other noise types. Option $A$ likely makes the relative transformations too unbalanced for the calibration, as seen from \cref{fig:transformation}. The data is generated with a ground vehicle in mind, so the transformations are heavily partial to the $x$, $y$, and $yaw$ components.

For added outliers and drift, the performance of the selection strategies is ordered from those providing the smallest transformations to those providing the largest. In case of DNLO, the outliers are also more prominent when using the short reference $B1$, and therefore easier to reject. For added drift, the differences between the different strategies are minute, apart from option $A$. The mixed noise case in \cref{fig:dnlo_mixed} provides the most interesting notion, where $B1$ is clearly best w.r.t. the relative errors, but not w.r.t. the absolute error, which is the true measure of calibration performance. This is noteworthy when selecting possible reference frames, as under normal operating conditions ${}_{s1}\mathbf{T}^{s2}_{gt}$ is unknown and one can only reason using the available relative error metrics. This means that the relative error metrics can not be used as a reliable selection criteria when choosing suitable reference frames.

\subsubsection{Comparing best performing algorithms}

The best performing variant, w.r.t. $e_{at}$~\cref{eq:eat}, of each algorithm are compared in \cref{fig:best} and follow a similar pattern as already discussed regarding DNLO. In the case of added Gaussian noise, preprocessing option $B10$ performs best for all algorithms, with the exception of Zhuang~\citep{Zhuang1994}. The likely cause is the general reasoning that larger transformations between the points are less susceptible to noise.

For added outliers, the \emph{de facto} preprocessing $B1$ again provides the best results for most algorithms. However, it is clear that the added outlier rejection of DNLO outperforms the others. The poor performance of Ali~\citep{Ali2019} is likely explained by the rotation error not directly affecting the cost. The very small rotation errors in the \emph{separable} algorithms Park~\citep{Park2020} and Taylor~\citep{Taylor2015}, in both added outliers and drift, are due to the simulated noise only affecting the Cartesian coordinates. As the \emph{separable} algorithms solve rotation first, it is unaffected by the noise. For added drift, there is again very little difference between the tested reference selection methods, apart from $A$, for any of the algorithms. Under the mixed noise conditions $B5$ performed best for all algorithms. This was to be expected, as it provides a balance between the short reference distance, favoring added outliers and drift, and the longer ones favoring added Gaussian noise. As with DNLO, upon selecting the possible reference frames, $B1$ appears better using the error metrics available.

\input{tables/kitti_keyframe_vsb1}
\input{tables/kitti_long_cam_vsb1}

\section{Experiments on KITTI data}

To validate the findings of the simulation tests with real data, we perform experiments on the KITTI dataset~\citep{Geiger2013}, which is widely used and publicly available. The dataset, collected by driving around residential areas, contains data from two stereo cameras, a Velodyne HDL-64E lidar, and a GPS/INS system. We selected two drive sequences from the dataset, \texttt{2011\_10\_03\_drive\_0027}, a longer sequence with over 4000 frames, and \texttt{2011\_09\_30\_drive\_0027}, a shorter sequence with around 1000 frames. The first was selected as it is also used in \citep{Taylor2015}, and the latter was randomly selected to validate that the hand-eye calibration method works with varying amounts of data.

As the ability to calibrate multimodal sensor-setups is one of the main draws of the demonstrated hand-eye calibration methods, we test both camera to lidar and camera to camera calibration. The calibration values provided in the dataset are used as ground truth values for computing the absolute error metrics.

\subsection{Trajectory generation}

To generate the trajectories, the KITTI raw data was first exported into ROS bag format to allow the use of existing ROS implementations of the chosen SLAM algorithms. ORB SLAM3~\citep{Campos2021} was used for stereo visual SLAM for both the grayscale and color stereo cameras available in the dataset, and HDL graph SLAM~\citep{Koide2019} for LIDAR-based SLAM, using the point clouds from Velodyne HDL-64E. The trajectories were generated based on the keyframe poses after the optimization step in both algorithms.

ORB SLAM3 is currently considered to be the state-of-the-art visual and visual-inertial SLAM system for stereo cameras. It is a bundle adjustment based SLAM that uses ORB features and descriptors matching based tracking in the visual front-end. Its multi-map system allows superior mid-term and long-term data association necessary for relocalization when the tracking is lost and loop closure detection. HDL graph SLAM is a pose graph-based method, where GICP based scan matching is used for generating odometry. It utilizes distance-based loop closure detection, where GICP is again used for scan matching, and loop closure matches as posegraph constraints.

\subsection{Results}

As with the simulation data, we focus on the best performing variant of each calibration algorithm. The calibration errors on camera to lidar calibration on the short residential data sequence are presented in \cref{tab:camLidar0930}. Zhuang~\citep{Zhuang1994}, while having the lowest relative errors, converges to a wrong solution. All other algorithms perform best when provided with larger relative transformations, $B5$ or $B10$.

Similarly, the best performing variants for the camera to camera calibration of the long residential sequence, presented in \cref{tab:camCam1003}, are those using the midlength steps in reference selection. For DNLO, the keyframe based selection $C5$ is marginally better than $B5$, which performs best for all other algorithms.

It is noteworthy, that the performance gained by choosing a suitable reference frame in most cases outweighs the difference between the different calibration algorithms. The tests further demonstrate that using the \emph{de facto} reference $i=j-1$ is not the best option, but future work is needed to study how to best determine suitable spacing to the reference frame.

\section{Conclusion}

We demonstrated that the type of noise present in the sensor trajectories affects which reference selection strategy performs best in motion-based hand-eye calibration for sensor extrinsics. Notably, the reference selection strategy typically used throughout previous studies was deemed to perform worse than others with realistic data. However, this can only be observed with a known ground truth transformation and the relative error metrics available when selecting the reference frames can not be used as reliable selection criteria. Further research is needed to determine how to best select the reference frames based on given trajectories. In most cases, the choice of reference frame selection had more impact on the calibration performance than the choice of the tested state-of-the-art calibration algorithm.

We also proposed a cost for nonlinear optimization to mitigate the effect of outliers in the calibration. The method was evaluated with respect to state-of-the-art, and either matched or outperformed the other tested algorithms in most noise conditions.

\vspace{6pt} 

\section*{Author contributions}
Conceptualization and methodology, T.V. and R.G.; investigation and data curation, T.V and B.G.; software, validation, formal analysis, visualization, and writing---original draft preparation, T.V.; writing---review and editing, T.V. and R.G.; supervision, project administration, and funding acquisition, R.G. All authors have read and agreed to the published version of the manuscript.

\section*{Funding}
This work was supported by AI Hub Tampere, funded by ERDF (EU Regional Development Funding), the Council of Tampere Region (Pirkanmaan liitto), FIMA (Forum for Intelligent Machines), Business Tampere, and Business Finland. This work was also supported by PEAMS (Platform Economy for Autonomous Mobile Machines Software Development), funded by Business Finland.

\section*{Data availability}
The data presented in this study and all software needed to reproduce the results are openly available through GitHub at \url{https://github.com/tau-alma/trajectory_calibration_experiments} and \url{https://github.com/tau-alma/trajectory_calibration}.

\section*{Conflicts of interest}
The authors declare no conflict of interest. The funders had no role in the design of the study; in the collection, analyses, or interpretation of data; in the writing of the manuscript; or in the decision to publish the~results.

\addtolength{\textheight}{-12.5cm}

\bibliography{bibliography}

\end{document}

%% file: tables/kitti_keyframe_vsb1.tex
\begin{table*}[!ht]
\centering
\caption{Camera to lidar calibration on KITTI \textnormal{\texttt{2011\_09\_30\_drive\_0027}}}
\label{tab:camLidar0930}
\begin{tabular}{lS[table-format=1.3]S[table-format=1.3]S[table-format=1.3]S[table-format=1.3]S[table-format=1.3]S[table-format=1.3]}
\toprule
{} & \multicolumn{2}{c}{Relative error} & \multicolumn{2}{c}{Absolute error} & \multicolumn{2}{c}{Improvement over B1} \\
{Method} & {$e_{rt}$ (m)} & {$e_{rR}$ (deg)} & {$e_{at}$ (m)} & {$e_{aR}$ (deg)} & {$e_{at}$ (m)} & {$e_{aR}$ (deg)} \\
\midrule
Ali B5 & 0.170 & 0.294 & 0.342 & 0.722 & 0.288 & -0.056 \\
DNL B5 & 0.170 & 0.293 & 0.342 & 0.721 & 0.288 & -0.052 \\
DNLO B10 & 0.271 & 0.488 & 0.202 & \bfseries 0.232 & 0.128 & 0.209 \\
Park B10 & 0.325 & 0.473 & \bfseries 0.183 & 0.849 & 0.435 & -0.102 \\
Taylor B10 & 0.325 & 0.473 & \bfseries 0.183 & 0.849 & 0.435 & -0.102 \\
Zhuang B1 & \bfseries 0.055 & \bfseries 0.152 & 1.063 & 2.899 & {-} & {-} \\
\bottomrule
\end{tabular}
\end{table*}

%% file: tables/kitti_long_cam_vsb1.tex
\begin{table*}[!ht]
\centering
\caption{Camera to camera calibration on KITTI \textnormal{\texttt{2011\_10\_03\_drive\_0027}}}
\label{tab:camCam1003}
\begin{tabular}{lS[table-format=1.3]S[table-format=1.3]S[table-format=1.3]S[table-format=1.3]S[table-format=1.3]S[table-format=1.3]}
\toprule
{} & \multicolumn{2}{c}{Relative error} & \multicolumn{2}{c}{Absolute error} & \multicolumn{2}{c}{Improvement over B1} \\
{Method} & {$e_{rt}$ (m)} & {$e_{rR}$ (deg)} & {$e_{at}$ (m)} & {$e_{aR}$ (deg)} & {$e_{at}$ (m)} & {$e_{aR}$ (deg)} \\
\midrule
Ali B5 & 0.155 & 0.181 & \bfseries 0.074 & 0.432 & 0.104 & 0.018 \\
DNL B5 & 0.155 & 0.181 & \bfseries 0.074 & 0.432 & 0.104 & 0.018 \\
DNLO C5 & \bfseries 0.076 & \bfseries 0.149 & 0.159 & \bfseries 0.345 & 0.034 & 0.071 \\
Park B5 & 0.157 & 0.181 & 0.078 & 0.351 & 0.111 & 0.075 \\
Taylor B5 & 0.157 & 0.181 & 0.078 & 0.351 & 0.111 & 0.075 \\
Zhuang B5 & 0.155 & 0.181 & \bfseries 0.074 & 0.400 & 0.106 & -0.081 \\
\bottomrule
\end{tabular}
\end{table*}